\documentclass[runningheads]{llncs}

\usepackage{url}
\usepackage{graphicx}
\usepackage{authblk}
\usepackage{wrapfig}
\usepackage{amssymb}
\usepackage{amsthm}
\usepackage{amsmath}
\usepackage{algorithm}
\usepackage{algorithmic}
\usepackage{subcaption}
\newcommand{\nrm}[1]{\ensuremath{\|#1\|_\mathcal{H}}}
\newcommand{\iprod}[2]{\ensuremath{\langle #1, #2 \rangle_\mathcal{H}}}
\newcommand{\bmu}{\ensuremath{\boldsymbol \mu}}

\newcommand\numberthis{\addtocounter{equation}{1}\tag{\theequation}}

\begin{document}
\title{The Kernelized Taylor Diagram}
%
%
\author{Kristoffer Wickstrøm\inst{1}\orcidID{0000-0003-1395-7154} \and
J. Emmanuel Johnson\inst{2}\orcidID{0000-0002-6739-0053} \and
Sigurd Løkse\inst{1}\orcidID{0000-0002-1953-4315} \and
Gustau Camps-Valls\inst{2}\orcidID{ 0000-0003-1683-2138} \and
Karl Øyvind Mikalsen\inst{1,4}\orcidID{0000-0003-4672-7865} \and
Michael Kampffmeyer\inst{1,3}\orcidID{0000-0002-7699-0405} \and
Robert Jenssen\inst{1,3}\orcidID{0000-0002-7496-8474}}
\authorrunning{K. Wickstrøm et al.}
%
\institute{UiT the Arctic University of Norway, Tromsø, Norway \and
Universitat de València, València, Spain \and
Norwegian Computing Center, Oslo, Norway \and 
University Hospital of North Norway, Tromsø, Norway}
\maketitle
\begin{abstract}
  This paper presents the kernelized Taylor diagram, a graphical framework for visualizing similarities between data populations. The kernelized Taylor diagram builds on the widely used Taylor diagram, which is used to visualize similarities between populations. However, the Taylor diagram has several limitations such as not capturing non-linear relationships and sensitivity to outliers. To address such limitations, we propose the kernelized Taylor diagram. Our proposed kernelized Taylor diagram is capable of visualizing similarities between populations with minimal assumptions of the data distributions. The kernelized Taylor diagram relates the maximum mean discrepancy and the kernel mean embedding in a single diagram, a construction that, to the best of our knowledge, have not been devised prior to this work. We believe that the kernelized Taylor diagram can be a valuable tool in data visualization.
\end{abstract}

\section{Introduction}
Clear and informative visualization of similarities between populations is a key component both in the development of methodology and in scientific publications. Depending on the particular use case, a wide range of techniques are available. One such visualization technique is the Taylor diagram (TD) \cite{Tdiagram}, which was devised to relate several statistical quantities and allow for comparison of numerous data points in a single diagram. The TD has been frequently used in numerous application, and particularly in climate sciences \cite{TDex1,TDex2}. However, the statistical quantities displayed in the TD does have some weaknesses that limit the usability of the diagram. For instance, one quantity in the diagram is the Pearson correlation coefficient, which only models linear relationship and can be sensitive to outliers. This curtails the TD, as many real-world applications use data with outliers and that are connected through non-linear relationships.

\begin{wrapfigure}{r}{0.5\linewidth}
    \centering
    \includegraphics[width=0.475\columnwidth]{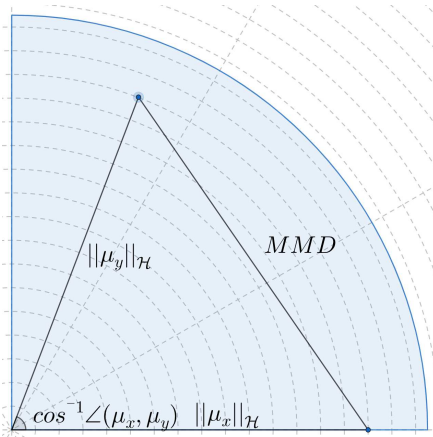}
    \caption{KTD: The radial distance from the origin to each point is proportional to the length of kernel mean embedding. The distance between the points is the maximum mean discrepancy.}
    \label{fig:KTDill}
\end{wrapfigure}

One of the most well-known and widely used approaches for measuring similarity in machine learning is through kernel methods \cite{KA,CKA}. At its core, a kernel function corresponds to a dot product in a high-dimensional feature space, where non-linear relationship between data in the input space can be linearly related in the new feature space. As long as the kernel is positive definite, the mapping to the feature space does not have to be computed explicitly. 

In this paper we propose the kernelized Taylor diagram (KTD), which is illustrated in Fig. \ref{fig:KTDill}. This diagram relates well-known quantities from the kernel literature, namely the maximum mean discrepancy (MMD) and the kernel mean embedding in a single figure. To the best of our knowledge, such a diagram has never been devised prior to this work. The KTD makes no assumptions on the distributions of the populations and can model a rich family of relationships between populations. The functionality of the proposed diagram is demonstrated on synthetic data. Code: \url{https://github.com/Wickstrom/KernelizedTaylorDiagram}.

\section{The Kernelized Taylor Diagram} \label{sec:KTD}

\paragraph{\textbf{Taylor diagram}} The TD was introduced as a tool that could relate several statistical quantities in a single figure \cite{Tdiagram}. It strength lies in the ability to compare numerous data points where it would otherwise be necessary to utilized several figures and/or tables. The theoretical starting point of the TD is the Pearson correlation coefficient $\rho$ and the root-mean-squared-error $E$ between two data points. \cite{Tdiagram} argued that neither are sufficient to capture potential similarities on their own, but in the aggregate the they are capable of detecting a wide range of differences between data points. Let $\mathbf{x}$ and $\mathbf{z}$ represent two $D$-dimensional vectors representing two data points. The correlation coefficient between $\mathbf{x}$ and $\mathbf{z}$ is defined as:

\begin{equation}\label{eq:CorrCoeff}
    \rho = \frac{1}{d}\sum\limits_{d=1}^D\frac{(x_d-\bar{x})(y_d-\bar{z})}{\sigma_x\sigma_y},
\end{equation}
where $\bar{x}$ and $\bar{y}$ are the mean values and $\sigma_x$ and $\sigma_y$ are the standard deviations. The root-mean-squared-error for mean centered data points is defined as:

\begin{align*}
    E^2 &= \mathbb{E}\Bigg[\frac{1}{d}\sum\limits_{d=1}^D\Big((x_d-\bar{x}) - (z_d-\bar{z})\Big)^2\Bigg] \\
    &=\underbrace{\frac{1}{D^2}\mathbb{E}\Big[\sum\limits_{d=1}^D(x_d-\bar{x})^2\Big]}_{\sigma_x^2}+\underbrace{\frac{1}{D^2}\mathbb{E}\Big[\sum\limits_{d=1}^D(y_d-\bar{y})^2\Big]}_{\sigma_y^2}-\underbrace{\frac{1}{D^2}\mathbb{E}\Big[\sum\limits_{d=1}^D(x_d-\bar{x})(y_d-\bar{y})\Big]}_{\sigma_{xy}} \\
    &= \sigma_x^2+\sigma_y^2-2\sigma_x\sigma_y\rho \numberthis \label{eq:SED}.
\end{align*}
The key point of the TD is recognize the relationship between the statistical quantities in Eq. \ref{eq:SED} and the law of cosines:

\begin{equation}\label{eq:lawofcos}
    c^2 = a^2+b^2-2ab\cos(\theta).
\end{equation}
Here, $a$ and $b$ are the lengths of two sides of a triangle with angle $\theta$ between each other and an opposite side of length $c$. The TD has seen widespread use in several domains such as in geophysical sciences \cite{TDex1,TDex2}. Nevertheless, the TD has some key weaknesses that limits it functionality in many practical applications. The Pearson correlation coefficient has a number of limitations \cite{Armstrong2019}. It can only model linear relationships \cite{Correa2013THEMI}, which can be restricting in many practical application. Also, the Pearson correlation coefficient is known be sensitive to outliers \cite{Armstrong2019}.

\paragraph{\textbf{The kernelized Taylor diagram}} To address such limitations, we propose the KTD, which uses well-know measures from the kernel literature to model similarities between populations. The starting point of the KTD is one of the most widely used distance measures in the kernel literature, namely the maximum mean discrepancy (MMD) \cite{MMD}, which measures the distance between two distributions where each distributions is represented by a mean embedding of the data. Let $X\sim P$ and $Y\sim Q$, and  $\bmu_x$ and $\bmu_y$ denoted the mean embedding vectors representing two distributions $P$ and $Q$. Then, the MMD is defined as the norm between the two embeddings in a reproducing kernel Hilbert space $\mathcal{H}$:

\begin{equation}\label{eq:MMD}
    \begin{aligned}
        MMD^2 &= \nrm{\bmu_x - \bmu_y}^2 \\
              &= \nrm{\bmu_x}^2 + \nrm{\bmu_y}^2
              - 2 \iprod{\bmu_x}{\bmu_y} \\
              &= \nrm{\bmu_x}^2 + \nrm{\bmu_y}^2 - 2 \nrm{\bmu_x}\nrm{\bmu_y}
              \frac{\iprod{\bmu_x}{\bmu_y}}{\nrm{\bmu_x}\nrm{\bmu_y}} \\
              &= \nrm{\bmu_x}^2 + \nrm{\bmu_y}^2 - 2 \nrm{\bmu_x}\nrm{\bmu_y}
              \cos\angle(\bmu_x, \bmu_y).
    \end{aligned}
\end{equation}
In general, the true data distributions are not known, so the mean embeddings are replaced by empirical mean embeddings that are estimated based on samples from each distribution:

\begin{equation}\label{eq:EmpiricalME}
    \hat{\bmu}_x = \frac{1}{N}\sum\limits_{n=1}^N \kappa (\mathbf{x}_n, \cdot),
\end{equation}
where $\kappa(\cdot, \cdot)$ is a positive definite kernel that measures similarity between data points. If the kernel is characteristic \cite{MMD}, MMD is a metric and is zero only if the two distributions are equal. \cite{NIPS20073a077244} showed that the well-known Gaussian kernel with kernel width $\sigma$, $G_\sigma (\mathbf{x}_i, \mathbf{x}_j)=\exp(||\mathbf{x}_i-\mathbf{x}_j||^2/2\sigma)$, is a characteristic kernel. Furthermore, MMD does not assume a particular distribution of the data, and can capture both non-linear and linear relationships between distributions.

Similarly as with the TD, we recognize the law of cosines in Eq. \ref{eq:MMD}. The mean embeddings of the two distributions are the side lengths of a triangle with angle $\cos\angle(\bmu_x, \bmu_y)$ between each other and an opposite side with length equal to the MMD between the distributions. The KTD is shown in Fig. \ref{fig:KTDill}.

The length of the mean embeddings indicate the distance from the origin to each point in the KTD. For the Gaussian kernel, the kernel mean embedding captures all moments of the data population \cite{kernleMeanEmb}. But it is not obvious how to interpret what information the kernel mean embeddings are illustrating in the diagram. However, the kernel mean embeddings can be related to uncertainty through the information potential (IP) from information theoretic learning \cite{Xu2010}, which allows for a similar interpretation of the KTD as the TD. That is, the kernel mean embeddings correspond to the $\sigma$ in Eq. \ref{eq:SED}. In most applications, the IP must be estimated from data. In information theoretic learning, the IP is often estimated through the quadratic IP estimator using a Gaussian kernel \cite{Xu2010}:

\begin{equation}\label{eq:IP}
    \hat{V}_{2, \sigma}(X) = \frac{1}{N^2}\sum\limits_{i, j}^N G_\sigma (\mathbf{x}_i, \mathbf{x}_j).
\end{equation}
Next, the squared norm terms in Eq. \ref{eq:MMD} can be expressed as:

\begin{equation}\label{eq:HSnorm}
    \nrm{\bmu_x}^2 = \frac{1}{N^2}\sum\limits_{i, j}^N \kappa (\mathbf{x}_i, \mathbf{x}_j).
\end{equation}
If the mean embeddings are calculated using a Gaussian kernel, Eq. \ref{eq:IP} and Eq. \ref{eq:HSnorm} are equivalent. Furthermore, the IP is related to entropy as follows:

\begin{equation}\label{eq:QuadEntropy}
    \hat{H}_2(X) = -\log(\hat{V}_{2, \sigma}(X)).
\end{equation}
Entropy measures the amount of information in a random variable, but can also be interpreted as a measure of uncertainty. High entropy indicates more variation in the data, while low entropy means that the data is clustered together. From Eq. \ref{eq:QuadEntropy} it is evident that when the information potential of $X$ is high and the entropy will be low, and the opposite when the information potential of $X$ is low. For the KTD, this means that random variables with a high value for the kernel mean embedding, and thus far from the origin, is associated with low uncertainty, and oppositely for a low value of the kernel mean embedding. This insight is important, as it allows us to relate concepts from the TD to the KTD.

\section{Experiments} \label{sec:Ex}

\subsection{Illustration on Synthetic Data}\label{sec:Synthetic}
To illustrate the functionality of the KTD we consider the case were the true distribution of the data is known and generate 1000 samples from 5 different populations. The reference distribution $X_{\text{ref}}$ is sampled from a standard normal distribution. The remaining populations are constructed as follows:

\begin{figure*}[htb]%
    \centering
    \begin{subfigure}{0.46\linewidth}
    \includegraphics[width=0.975\linewidth]{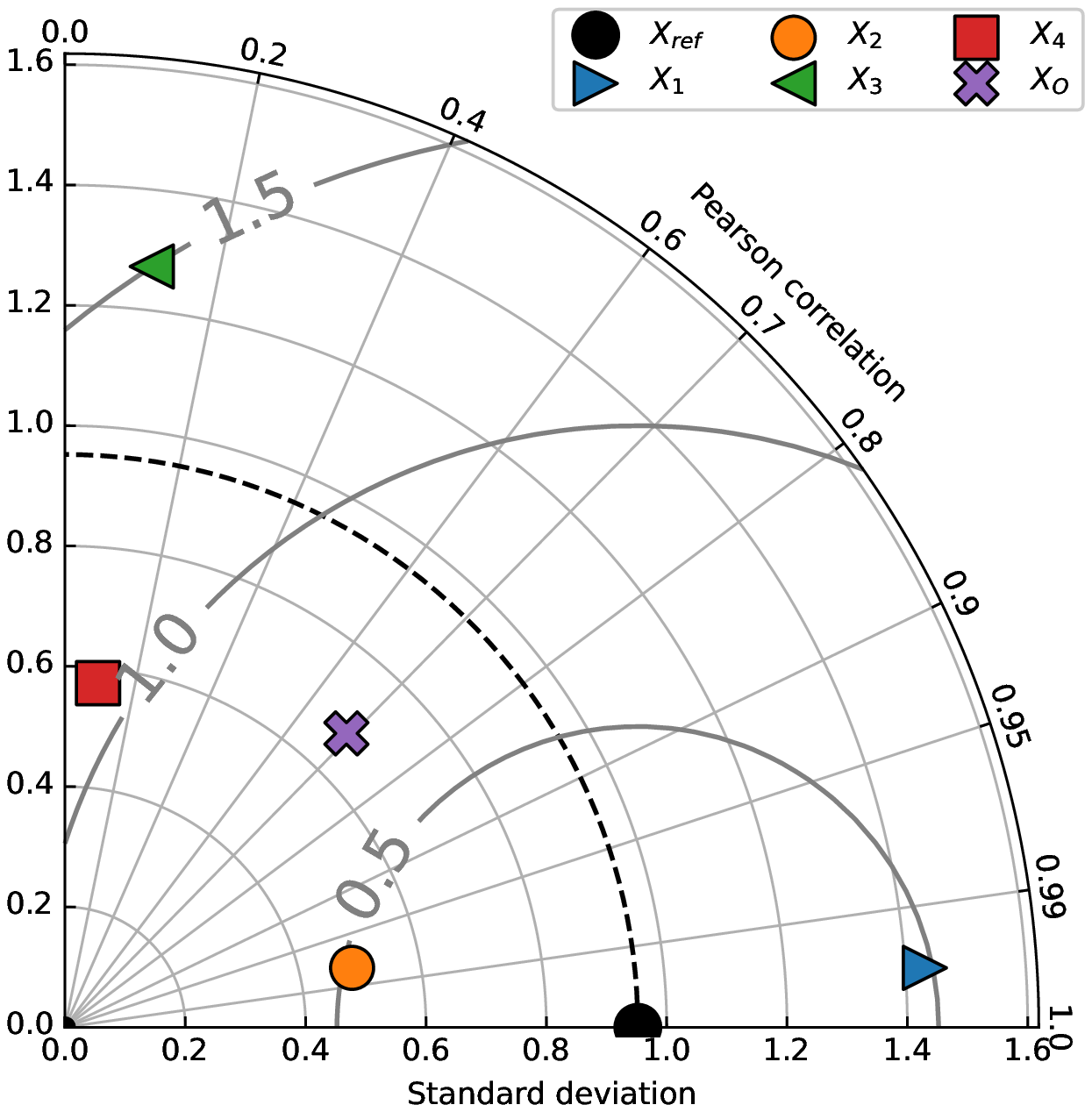}
    \caption{Taylor diagram}\label{fig:SyntheticPearson}
    \end{subfigure}
    \qquad
    \begin{subfigure}{0.46\linewidth}
    \includegraphics[width=0.975\linewidth]{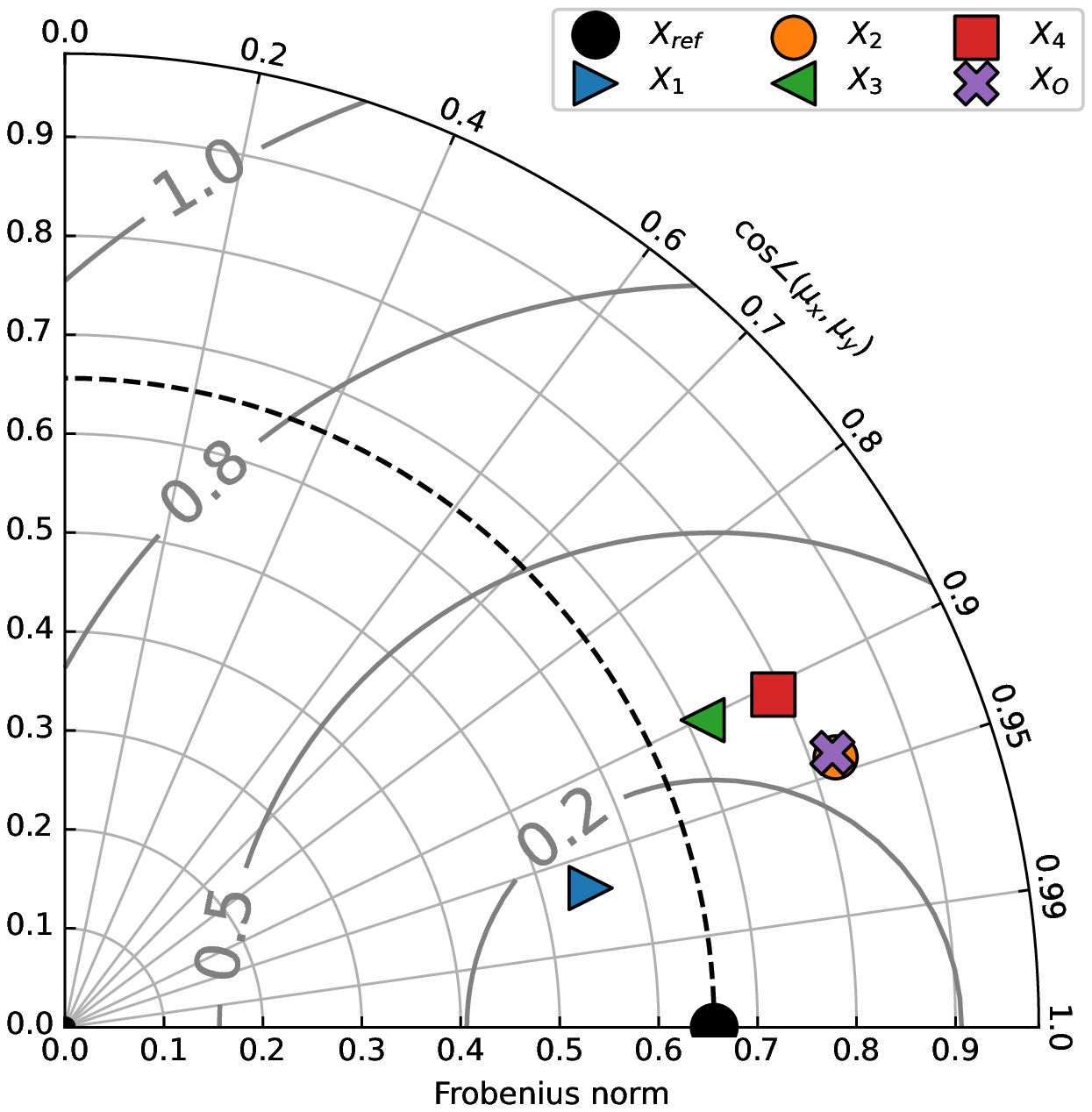}
    \caption{Kernelized Taylor diagram}\label{fig:SyntheticKTD}
    \end{subfigure}
    \caption{Comparison of TD with the KTD on the data described in Sec. \ref{sec:Synthetic}. The experiment illustrates how the TD is not able to capture non-linear dependencies and is sensitive to outliers, when compared with the proposed KTD.}%
    \label{fig:SyntheticData}%
\end{figure*}

\begin{align*}
    X_1 &\sim 2X_{\text{ref}}+\epsilon,
    \hspace{0.25cm} 
    X_2 \sim \frac{X_{\text{ref}}}{2}+\epsilon,
    X_3 \sim X_{\text{ref}}^2+\epsilon, \\
    \hspace{0.25cm}
    X_4 &\sim X_{\text{ref}}\sin(X_{\text{ref}})+\epsilon, X_O \sim \frac{X_{\text{ref}}}{2}+\epsilon \text{   (with outliers),}
\end{align*}
where $\epsilon\sim \mathcal{N}(0, 0.01)$. Population $X_1$ and $X_2$ are chosen to represent a linear relationship to the reference distribution, but with different scaling such that the standard deviation is different compared to the reference. Population $X_3$ and $X_4$ are chosen to represent a non-linear relationship with the reference. Lastly, $X_0$ is chosen to also have a linea relationship with the reference, but with two outliers added to the population. These two outliers are samples from $\mathcal{N}(10, 1)$.

Fig. \ref{fig:SyntheticPearson} displays the TD for these populations in relation to the reference distribution, while Fig. \ref{fig:SyntheticKTD} shows the KTD. First, we consider Fig. \ref{fig:SyntheticPearson}. Note that $X_1$ and $X_2$ both have a high similarity with the reference but with different length from the origin as a result of the difference in standard deviation. Next, both $X_3$ and $X_4$ are indicated as having low similarity with the reference, which is expected since the relationship is non-linear. Lastly, $X_O$, which is almost identical to $X_2$ except for two outliers, shows a much lower similarity score. This illustrates how sensitive the TD can be to outliers.

In Fig. \ref{fig:SyntheticKTD}, $X_1$ and $X_2$ also shows a related and high similarity score. However, note that compared to Fig. \ref{fig:SyntheticPearson}, the distance to the origin have been changed, which is explained through the connection to the information potential described in Sec. \ref{sec:KTD}. Next, both $X_3$ and $X_4$ are now indicated to have a high similarity with the reference, which illustrates that the KTD is capable of capturing non-linear similarities. Lastly, $X_2$ and $X_O$ are located at almost the same point in the diagram, which shows that the KTD is robust against outliers in the data.

\section{Conclusion}
In this article we proposed the KTD, which relates well-known quantities from the kernel literature in a single diagram. To the best of our knowledge, such a diagram has not been devised previously. Our proposed diagram addresses some key limitation in the widely used TD, such as modeling non-linear relationships and outliers in the data. In future works, we intend to examine the usability of the diagram on real-world data such as in climate applications. We believe that the KTD can be a useful tool in many machine learning applications.

\bibliographystyle{splncs04}
\bibliography{bibliography}

\end{document}